\documentclass[runningheads]{llncs}
\usepackage{graphicx}
\usepackage{comment}
\usepackage{amsmath,amssymb} %
\usepackage{color}
\usepackage{subcaption} %
\usepackage{booktabs} %
\usepackage{caption}  %
\usepackage{textcomp} %
\usepackage{wrapfig}
\usepackage{paralist}
\usepackage{xspace}
\usepackage{hyperref}

\usepackage{printlen} %

\DeclareMathOperator*{\argmin}{argmin~}

\makeatletter
\newcommand{\printfnsymbol}[1]{%
        \textsuperscript{\@fnsymbol{#1}}%
}
\makeatother

\begin{document}
\pagestyle{headings}
\mainmatter
\def\ECCVSubNumber{5315}  %
\pagestyle{headings}

\title{Category Level Object Pose Estimation via Neural Analysis-by-Synthesis} %

\author{Xu Chen\inst{1,3}\thanks{Equal contribution.} \and
Zijian Dong\inst{1}\printfnsymbol{1} \and
Jie Song\inst{1} \and
Andreas Geiger \inst{2,4} \and 
Otmar Hilliges \inst{1}}
\authorrunning{X. Chen et al.}
\institute{ETH Z\"{u}rich \and
University of T\"{u}bingen \and
Max Planck ETH Center for Learning Systems \and
Max Planck Institute for Intelligent Systems, T\"{u}bingen
}
\maketitle

\begin{abstract}
Many object pose estimation algorithms rely on the analysis-by-synthesis framework which requires explicit representations of individual object instances. In this paper we combine a gradient-based fitting procedure with a parametric neural image synthesis module that is capable of implicitly representing the appearance, shape and pose of entire object categories, thus rendering the need for explicit CAD models per object instance unnecessary. The image synthesis network is designed to efficiently span the pose configuration space so that model capacity can be used to capture the shape and local appearance (i.e., texture) variations jointly. At inference time the synthesized images are compared to the target via an appearance based loss and the error signal is backpropagated through the network to the input parameters. Keeping the network parameters fixed, this allows for iterative optimization of the object pose, shape and appearance in a joint manner and we experimentally show that the method can recover orientation of objects with high accuracy from 2D images alone. When provided with depth measurements, to overcome scale ambiguities, the method can accurately recover the full 6DOF pose successfully. 
\end{abstract}

\section{Introduction}

Estimating the 3D pose
of objects from 2D images alone is a long-standing problem in computer vision and has many down-stream applications such as in robotics, autonomous driving and human-computer interaction. One popular class of solutions to this problem is based on the analysis-by-synthesis approach.
The key idea of analysis-by-synthesis is to leverage a forward model (e.g., graphics pipeline) to generate different images corresponding to possible geometric and semantic states of the world.
Subsequently, the candidate that best agrees with the measured visual evidence is selected.
In the context of object pose estimation, the visual evidence may comprise RGB images \cite{li2018deepim,manhardt2018deep}, depth measurements \cite{besl1992method,zeng2017multi} or  features extracted using deep networks such as keypoints \cite{manhardt2018deep,tekin2018real,oberweger2018making,hu2019segmentation,peng2019pvnet} or dense correspondence maps  \cite{li2019cdpn,park2019pix2pose,zakharov2019dpod}. 
While such algorithms can successfully recover the object pose, a major limiting factor is the requirement to i) know which object is processed and ii) to have access to an explicit representation of the object for example, in the form of a 3D CAD model.

Embracing this challenge, in this paper we propose an algorithm that can be categorized as analysis-by-synthesis but overcomes the requirement of a known, explicit instance representation.
At the core of our method lies a synthesis module that is based on recent advancements in the area of deep learning based image synthesis.
More specifically, we train a pose-aware neural network to predict 2D images of objects with desired poses, shapes and appearances.
In contrast to traditional static object representations, neural  representations are able to jointly describe a large variety of instances, thereby extrapolating beyond the training set and allowing for continuous manipulation of object appearance and shape.
In addition, our network is designed to efficiently span the space of 3D transformations, thus enabling image synthesis of objects in unseen poses. 

After training, the synthesis module is leveraged in a gradient-based model fitting algorithm to jointly recover pose, shape and appearance of unseen object instances from single RGB or RGB-D images.
Since our neural image synthesis module takes initial pose and latent codes (e.g., extracted from an image) as input, it can be used as a drop-in replacement in existing optimization-based model fitting frameworks.
Given the initial pose and latent code, we generate an image and compare it with the target RGB image via an appearance based loss.
The discrepancy in appearance produces error gradients which back-propagate through the network to the pose parameters and the latent code.
Note that instead of updating the network parameters as during training time, %
we now keep the network weights fixed and instead update the appearance, shape and pose parameters.
We repeat this procedure until convergence.

We evaluate our method on a publicly available real-world dataset on the task of category-level object pose estimation. Using RGB images, our method is able to estimate the 3D object orientation with an accuracy on par with and sometimes even better than a state-of-the-art method which leverages RGB-D input.
As demonstrated in previous work \cite{xiang2017posecnn},  RGB-only methods suffer strongly from inherent scale ambiguities.
Therefore, we also investigate an RGB-D version of our model which faithfully recovers both 3D translation and orientation.
We systematically study algorithmic design choices and the hyper-parameters involved during both image generation and model fitting. In summary:
\begin{itemize}
    \item We integrate a neural synthesis module into an optimization based model fitting framework to simultaneously recover object pose, shape and appearance from a single RGB or RGB-D image.
    \item This module is implemented as a deep network that can generate images of objects with control over poses and variations in shapes and appearances.
    \item Experiments show that our generative model reaches parity with and sometimes outperforms a strong RGB-D baseline. Furthermore, it significantly outperforms discriminative pose regression.\footnote{Project homepage: \url{ait.ethz.ch/projects/2020/neural-object-fitting}}%
\end{itemize}
\section{Related Work}

\subsection{Object Pose Estimation}

Given its practical importance, there is a large body of work focusing on object pose estimation. The state-of-the-art can be broadly categorized into template matching and regression techniques. Template matching techniques align 3D CAD models to observed 3D point clouds \cite{besl1992method,zeng2017multi}, images \cite{li2018deepim,manhardt2018deep}, learned keypoints \cite{manhardt2018deep,tekin2018real,oberweger2018making,hu2019segmentation,peng2019pvnet} or correspondence features \cite{li2019cdpn,park2019pix2pose,zakharov2019dpod}.
In contrast, \cite{rad2017bb8,kehl2017ssd,xiang2017posecnn} tackle object pose estimation as a classification or regression problem. However, to achieve high accuracy, these methods typically require template-based refinement,~e.g., using ICP \cite{besl1992method}. While yielding impressive results, all aforementioned methods require access to an instance specific 3D CAD model of the object, both during training and test time. This greatly limits their applicability since storing and comparing to all possible 3D CAD models at test time is impractical in many situations. Moreover, capturing high-fidelity and complete 3D models is often difficult and for some applications even impossible. 

Only recently, researchers started the attempt to address object pose estimation without requiring access to instance-specific 3D object models at test time. NOCS \cite{wang2019normalized} proposes to tackle this problem by learning to reconstruct the 3D object model in a canonical coordinate frame from RGB images and then align the reconstruction to depth measurements. They train their reconstruction network using objects from the same categories, which is expected to generalize to unseen instances within the same category at test time. While our method also learns to solve the task from objects within the same category, our method is a fully generative approach which simultaneously recovers object pose, shape and appearance.
Thus, it allows for directly synthesizing the object appearance, eliminating the intermediate step of reconstructing the object in 3D. LatentFusion \cite{park2019latentfusion} proposes a 3D latent space based object representation for unseen object pose estimation. In contrast to ours, it requires multi-view imagery of the test object to form the latent space and depth measurements at test time. 
In contrast to both NOCS \cite{wang2019normalized} and LatentFusion \cite{park2019latentfusion}, our model enables 3D object pose estimation from a single RGB image as input.
\subsection{Pose Dependent Image Generation}
Pose or viewpoint dependent image generation has been studied in two settings. One line of work focuses on synthesizing novel views for a given source image by directly generating pixels \cite{kulkarni2015deep,TDB16a,olszewski2019tbn,sitzmann2019deepvoxels} or by warping pixels from source to target view \cite{zhou2016view,penner2017soft,tvsn_cvpr2017,sun2018multiview,chen2019mono}. While such techniques can be used to render objects in different poses, the object appearance and shape is controlled by the source image which cannot be optimized.

Another line of work tackles the problem of disentangled image generation \cite{chen2016infogan,higgins2017beta,locatello2019challenging,shen2019interpreting,gansteerability}, considering object pose as one factor among many. Recent works \cite{nguyen2019hologan} achieve appealing results on viewpoint/pose disentanglement using a 3D latent space. While all mentioned methods are able to generate objects in different poses, shape and appearances, the pose cannot be controlled precisely (e.g., rotation by a set amount of degrees), rendering their application to absolute object pose estimation tasks difficult. Inspired by \cite{nguyen2019hologan}, our network also adopts a 3D latent space. However, we utilize the model in a supervised fashion to integrate precise absolute pose knowledge into the latent representation during training. This is achieved by integrating the 3D latent space with a conditional VAE framework. In contrast to \cite{sitzmann2019deepvoxels} which also utilizes a 3D latent space, our model can jointly represent multiple instances of an object category and to generalize to unseen instances. Similar and concurrent to ours, \cite{mustikovelaCVPR20} uses pose-aware image generation for viewpoint estimation, but in a discriminative manner.

\begin{figure}[t]
    \centering
    \includegraphics[width=\textwidth]{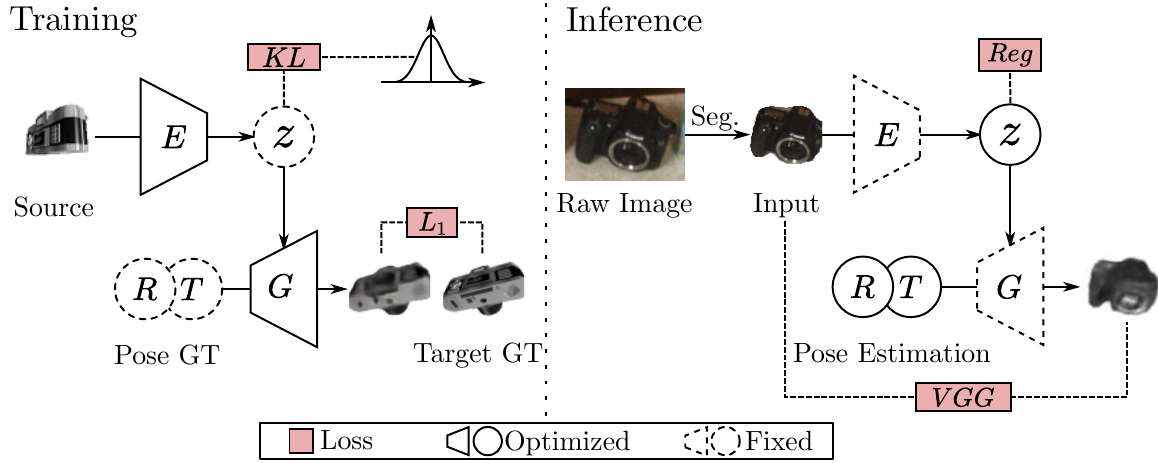}
    \caption{\textbf{Method overview.}
    We leverage a learned pose-aware image generator $G$ for object pose estimation.
     \emph{Training:} The generator is trained in the VAE framework by leveraging multi-view images of synthetic objects. We minimize the reconstruction loss between generated and ground-truth images of known orientation together with the KL divergence. %
     After training, the generator can produce images that faithfully reflect the (latent) appearance and desired pose.
    \emph{Inference:} To estimate object pose from a segmented real image as input, our method iteratively optimizes the object pose and shape, minimizing the perceptual loss between the input and the generated image while keeping the weights of the trained networks fixed.   }
    \label{fig:overview}
\end{figure}
\subsection{3D Representations for Objects}
Several works have addressed the problem of generating 3D geometric representations including meshes \cite{loper2015smpl,tan2018variational,gao2019sdm}, point sets \cite{achlioptas2017learning,yang2019pointflow,lin2018learning}, voxels \cite{wu2016learning,brock2016generative} and implicit functions \cite{park2019deepsdf,mescheder2019occupancy,chen2019learning,sitzmann2019scene}. While these generative models are also able to represent objects at category level and could theoretically be used for category-level object fitting in combination with a differential rendering algorithm, all aforementioned techniques only consider geometry, but not the appearance. As a result depth measurements are required and the rich information underlying in the object's appearance is discarded. In contrast, our method allows for leveraging appearance information and does not require depth maps as input. While \cite{VON} is able to generate textured objects, it is limited by its reliance on 3D meshes. In contrast, we completely forego the intermediate geometry estimation task, instead focusing on pose-conditioned appearance generation.
\subsection{Latent Space Optimization}
The idea of updating latent representations by iterative energy minimization has been exploited for other tasks. CodeSLAM \cite{bloesch2018codeslam} learns a latent representation of depth maps, optimizing the latent representation instead of per-pixel depth values during bundle adjustment. GANFit \cite{gecer2019ganfit} represents texture maps with GANs and jointly fits the latent code and 3DMM parameters to face images. \cite{abdal2019image2stylegan,gu2019image,shen2019interpreting,bau2019semantic} embeds natural images into the latent space of GANs by iteratively minimizing the image reconstruction error for image editing. Our method is inspired by these works, but targets a different application where latent appearance codes and geometric parameters such as poses must be optimized jointly.
\newcommand{\pose}{T}
\newcommand{\code}{z}
\newcommand{\network}{3DVAE }
\section{Method}
We propose a neural analysis-by-synthesis approach to category-level object pose estimation. Leveraging a learned image synthesis module, our approach is able to recover the 3D pose of an object from a single RGB or RGB-D image without requiring access to instance-specific 3D CAD models. 

Fig.~\ref{fig:overview} gives an overview of our approach. We first train a pose-aware image generator $G$ with multi-view images of synthetic objects from the ShapeNet \cite{chang2015shapenet} dataset, which is able to generate object images $\hat{I}=G(R,T,z)$ that faithfully reflect the input object pose ($R,T$) and appearance code $z$. At inference time with a segmented image $I$ as input, our method estimates object pose by iteratively optimizing the object pose and shape to minimize the discrepancy between the input image and the synthesized image $G(R,T,z$). 
\begin{figure}[t]
    \centering
    \includegraphics[width=\textwidth]{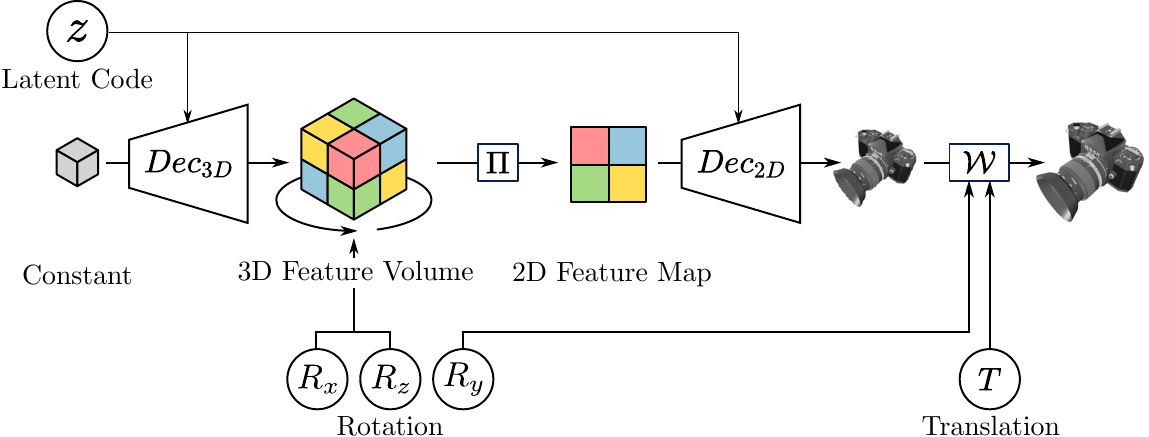}
    \caption{\textbf{Pose-aware image generator.} To generate an image in the desired pose $(R,T)$, the out-of-plane rotation $R_xR_y$ is first applied to the 3D feature volume, and then the 2D projection of the feature volume is decoded into an image. Subsequently, this image undergoes a 2D similarity transformation derived from the translation $T$ and in-plane rotation $R_z$ to form the final output. The latent code $z$ is injected into the 3D feature volume generator $Dec_{3D}$ and the 2D decoder $Dec_{2D}$ via adaIN to control the variation in shapes and appearances.}
    \label{fig:network}
\end{figure}
\subsection{Pose-aware Image Generator}

To generate images of different instances of a given category, we seek to generate images with significant but controllable shape and appearance variation. We encode shape and appearance via latent variable $z$, and the desired 6 DoF object pose, comprising 3D rotation $R=R_xR_yR_z$ and 3D translation $T=[t_x\ t_y\ t_z]$, via $(R,T)$. The $z$-axis is defined to align with the principal axis of the camera and the $y$-axis points upwards. 
For efficiency and to increase the capacity of our model in terms of representing the large variability of shapes and appearances, we decouple the image generation pipeline into two stages.
First, we observe that 3D translations $T$ and in-plane rotations $R_{z}$ can be modeled using 2D operations and thus do not need to be learned.
Therefore, we constrain our network $G_{3D}$ to generate only images with out-of-plane rotations $\hat{I}_{rot}= G_{3D}(R_x,R_y,z)$, i.e., elevation $R_{x}$ and azimuth $R_{y}$. The remaining transformations are modeled using 2D image warping operations $\mathcal{W}$ derived from 3D translations $T$ and in-plane rotation $R_z$. The full generation process is defined as $G=\mathcal{W}\circ G_{3D}$. 
\subsubsection{Appearance and 3D Rotation.} 
In order to generate images of objects in diverse appearances, shapes and poses, we adopt a 3D style-based image generation network similar to the one proposed in \cite{nguyen2019hologan} as illustrated in Fig.~\ref{fig:network}.
This network combines a 3D feature volume which faithfully captures 3D rotations with a style-based generator \cite{karras2019style}.
This enables our model to disentangle global appearance variation from geometric factors such as pose and shape. The 3D style-based image generation network consists of four main steps: i) generating the 3D feature volume, ii) transforming the feature volume based on the pose, iii) projecting the 3D volume into a 2D feature map, and iv) decoding the feature map into the predicted image $\hat{I}$.
Both, the 3D generator and the 2D generator are conditioned on the latent code $z$ via adaptive instance normalization \cite{huang2017arbitrary} to model the variance in shape and appearance respectively. The object orientation $R$ controls the transformation that is applied to the 3D feature volume. 
\subsubsection{Translation and 2D Rotation.} 
While the 3D style-based decoder can in principle cover the entire space of 6 DoF poses, 
the resulting model would require a very large capacity and dataset to be trained on.
We therefore constrain the decoder to out-of-plane rotations and handle all remaining transformation using a similarity 2D transformation. The warping field is given by:
\begin{align}
\mathcal{W}(T, R_z):  \begin{bmatrix} u\\v \end{bmatrix} \mapsto \frac{f}{t_z} \cdot \left(R_z \begin{bmatrix} u\\v \end{bmatrix} + \begin{bmatrix} t_x\\t_y \end{bmatrix}\right)
\end{align}
We use $\mathcal{W}$ to warp the generated image, yielding the final image $\hat{I}$: 
\begin{align}
\hat{I} = G(R,T,z) = \mathcal{W}(T, R_z) \circ G_{3D}(R_x,R_y,z)
\label{equ:final}
\end{align}%
\subsection{Training}
We train our image generator in a conditional VAE framework as illustrated in Fig.~\ref{fig:overview} in order to achieve precise pose control over the generated image. A VAE is an auto-encoder trained by minimizing a reconstruction term and the KL divergence between the latent space distribution and a normalized Gaussian. We use our 3D style-based image generation network as decoder and a standard CNN as encoder.

At each training iteration, the encoder first extracts the latent code from an image of a randomly chosen training object. Then the 3D image generation network takes this latent code together with the desired pose as input to generate an image $\hat{I}$ of the chosen object in the \emph{desired} pose. The encoder and decoder are jointly trained by minimizing the reconstruction loss between the generated image and the ground-truth, regularized via the KL divergence:
\begin{align}
    \mathcal{L} = \|I - \hat{I}\|_1  + \lambda_{KL}~D_{KL}
\end{align}
where $\lambda_{KL}$ weights the regularization term and is set to $1e^{-2}$ in our experiments. The required training data, namely images of objects in difference poses and the corresponding pose label, is obtained by rendering synthetic objects from the ShapeNet dataset \cite{chang2015shapenet}. Since translation and 2D rotation is modelled by similarity transformation which does not require training, we only generate training samples with out-of-plane rotations for the sake of training efficiency.

\subsection{Object Pose Estimation}

The trained pose-aware image generator can render objects in various shapes, appearances and poses.
Since the forward process is differentiable, we can solve the inverse problem of recovering its pose, shape and appearance by iteratively refining the network inputs (i.e., the pose parameters and the latent code), so that the discrepancy between the generated and the target image is minimized: 
\begin{align}
    R^*,T^*,z^* = \argmin_{R,T,z}~E(I, R, T, z)
\end{align}
In the following sections, we will discuss the choice of energy function $E$, our initialization strategy and the optimizer in detail. Note that we assume that the object is segmented from the background which can be achieved using off-the-shelf image segmentation networks such as Mask-RCNN \cite{he2017mask}.

\subsubsection{Energy Function.}

We require a distance function $d$ that measures the discrepancy between two images. Common choices include the $L_1$ and $L_2$ per-pixel differences, the Structural Similarity Index Metric (SSIM) \cite{wang2004image} and  the perceptual loss \cite{johnson2016perceptual} which computes the $L_2$ difference in deep feature space. To gain robustness with respect to domain shifts we adopt the perceptual loss as distance function and experimentally validate that it yields the best results.

Without further regularization, we found that the model may converge to degenerate solutions by pushing the latent code beyond the valid domain. To avoid such undesirable solutions, we penalize the distance to the origin of the latent space. Due to the KL divergence term used during training, codes near the origin are more likely to be useful for the decoder.
Our final energy function is then given by:
\begin{align}
    E(I, R, T, z) = \| F_{vgg}(I) - F_{vgg}(\hat{I}) \|_2 + \| z \|_2,
\end{align}
where $F_{vgg}$ is a VGG network \cite{Simonyan15} pre-trained on ImageNet \cite{deng2009imagenet} for deep feature extraction and $\hat{I}$ is the image generated based on $R$, $T$ and $z$ according to Eq.~\ref{equ:final}.

\subsubsection{Initialization Strategy.} Since the above energy function is non-convex, gradient-based optimization techniques are prone to local minima. Therefore, we start the optimization process from multiple different initial states which we execute in parallel for efficiency. The optimal solution results in an image that aligns with the target image as well as possible, with any differences in pose or appearance leading to an increase in the reconstruction error. We thus choose the best solution by evaluating the optimization objective. 

We sample the initial poses uniformly from the space of valid poses.
As the latent dimension is high, random sampling can be inefficient due to the curse of dimensionality.
We therefore leverage our (jointly trained) encoder to obtain mean and variance for the conditional distribution of the latent code and sample from the corresponding Gaussian distribution.

\section{Evaluation}

We first compare our approach with the state-of-the-art category-level object pose estimation method NOCS \cite{wang2019normalized}. To validate the effectiveness of our method, we then also compare our method with several other baselines for category-level object model fitting. Furthermore, we systematically vary the architecture and hyper-parameters of our image generation network to analyze their influence on the fitting result. We also study the influence of the hyper-parameters for optimization. Finally, we evaluate the robustness of our generative model with respect to domain shifts, comparing it to a discriminative model.

\subsection{Comparison with state-of-the-art}

\subsubsection{Baseline.} To the best of our knowledge, NOCS \cite{wang2019normalized} is the only method for category-level object pose estimation. It uses both synthetic data generated from ShapeNet CAD models, and real data to train a network that is able to reconstruct objects from a RGB image in a canonical coordinate frame. Subsequently, the object pose is recovered by aligning the reconstruction to the depth map. NOCS uses both simulated (CAMERA) and real data (REAL275) for training. Their simulated data is generated by compositing synthetic images from \cite{chang2015shapenet} onto real-world tables. REAL275 contains real RGB-D images (4300 for training and 2750 for testing), capturing 42 real object instances of 6 categories (camera, can, bottle, bowl, laptop and mug) in 18 different scenes. 

Our method is trained with the same set of synthetic objects used in the CAMERA dataset. However our method does not require superimposing the objects onto real contexts. More importantly, in contrast to NOCS, our method does not require real images and their pose annotations for training. Note that we rely only on cheap to acquire 2D annotations of real images to fine-tune the object segmentation network (we use the REAL275 dataset for this purpose).

Since there is no public method capable of category-level object pose estimation from RGB images alone, we introduce a simple baseline consisting of a VGG16 \cite{Simonyan15} network that regresses object orientations directly from 2D images, trained on the same synthetic data as ours. Note that this is a fair comparison since the baseline and our method use the same training and test data.

\subsubsection{Metrics.} We follow the evaluation protocol defined by NOCS \cite{wang2019normalized}. NOCS reports average precision which considers object detection, classification and pose estimation accuracy. They set a detection threshold of 10\% bounding box overlap between prediction and
ground truth to ensure that most objects are included in the
evaluation. Next, they compute the average precision for different thresholds of rotation and translation error as performance indicator.
Although our method assumes that the target image does not contain background and we rely on a off-the shelf 2D image segmentation network for object detection, we follow the same protocol for a fair comparison with NOCS. To remove the influence of varying detection accuracy from the comparison, we use the trained Mask-RCNN network from the NOCS Github repository.

The error in rotation $e_{R}$ and translation $e_{t}$ is defined by:
\begin{align}
&e_R = arccos\frac{Tr(\Tilde{R} \cdot R^T )-1}{2} \text{ and } e_t = \|\Tilde{t} - t \|_2
\end{align}
where $Tr$ represents the trace of the matrix. For symmetric object categories (bottle, bowl, and can), we allow the predicted 3D bounding box to freely rotate around the object’s vertical axis with no penalty, as done in \cite{wang2019normalized}.

\subsubsection{Results: Translation.}

\begin{figure}[t!]
    \centering
    \includegraphics[width=\textwidth]{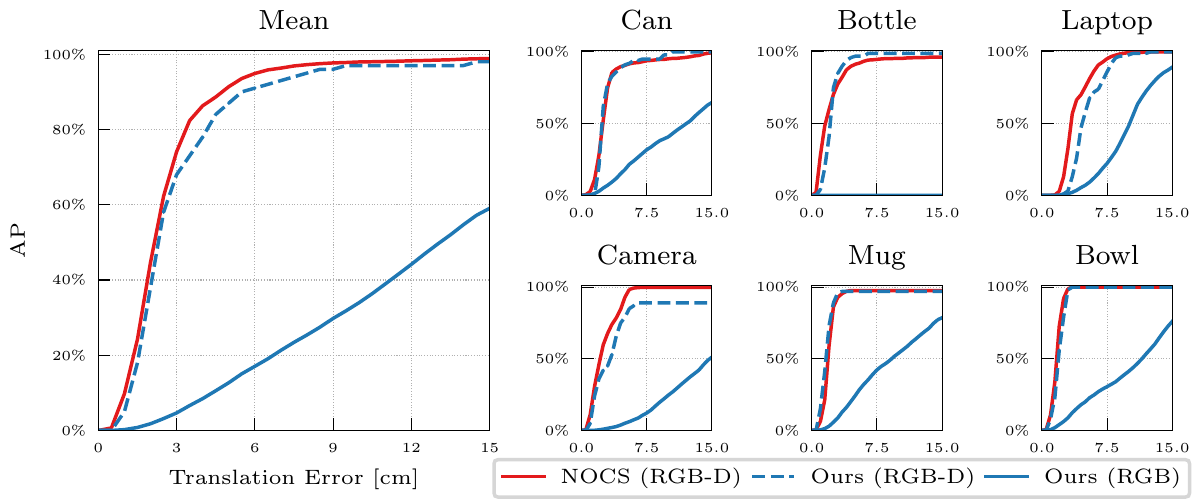}
    \caption{ \textbf{Comparison with NOCS: Translation.} Average precision at different translation thresholds. When using depth, our method achieves comparable results to the RGB-D method. When using RGB only, our method yields higher errors due to scale ambiguities.}
    \label{fig:exp_nocs_trans}
\end{figure}

We first evaluate the translation accuracy, summarized in Fig.~\ref{fig:exp_nocs_trans}. When using depth as input, both models perform comparable. We simply treat depth as an additional color channel during fitting and the translation parameter $T_z$ is directly added to the generated depth map. This version is trained on synthetic RGB-D data instead of RGB data. Without using depth at training and test time, our RGB only version does not achieve the same accuracy. This is however to be expected due to the inherent scale-ambiguity of 2D observations. We remark that similar observations have been made in the instance-level pose estimation literature \cite{xiang2017posecnn}.

\subsubsection{Results: Orientation.}

\begin{figure}[t]
    \centering
    \includegraphics[width=\textwidth]{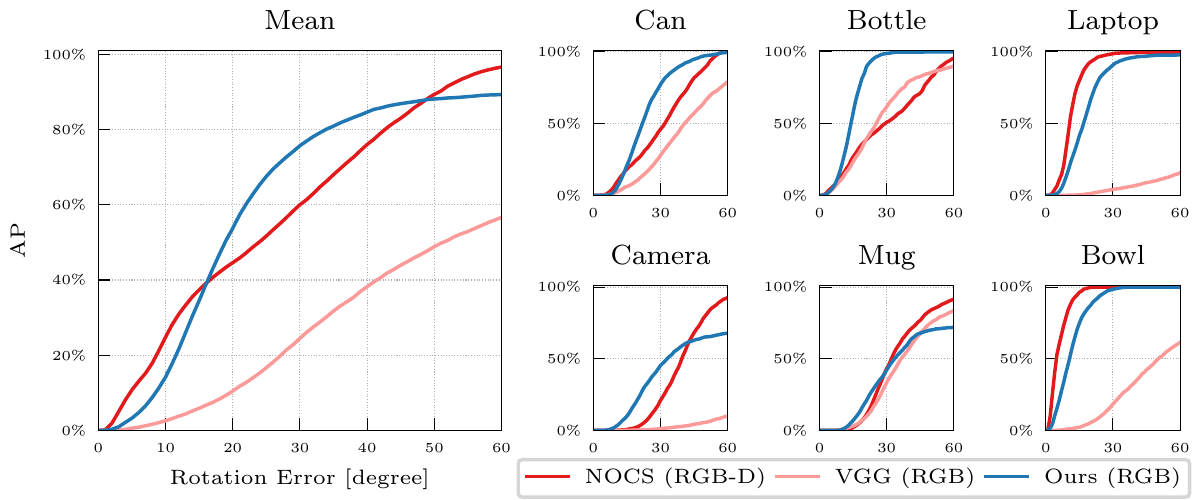}
    \caption{\textbf{Comparison with NOCS: Orientation.} Average precision at different  rotation error thresholds. Using only RGB as input, on average we achieve results on par or better than the NOCS baseline which uses RGB-D input and real training images with pose annotations. Our method can handle objects with complex geometry (camera) and textures (can) better than NOCS. See Fig. \ref{fig:exp_nocs_visual}.}
    \label{fig:exp_nocs_rot}
\end{figure}

\begin{figure}[t]
    \centering
    \includegraphics[width=\textwidth]{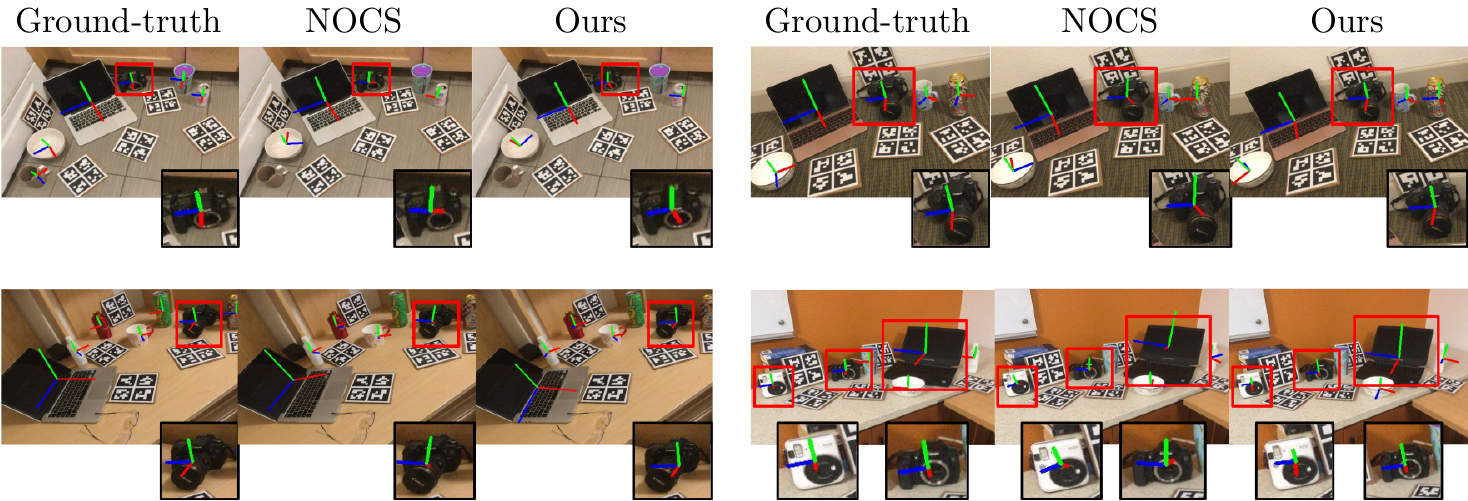}
    \caption{\textbf{Qualitative comparison to NOCS.} Our method can handle geometrically complex objects such as cameras better. Objects in insets.}
    \label{fig:exp_nocs_visual}
\end{figure}

The AP curve for our rotation estimation experiment is shown in Fig.~\ref{fig:exp_nocs_rot}.
Despite using only RGB inputs, on average we achieve results on par or better than the privileged NOCS baseline which uses RGB and depth as well as real-images with paired pose annotations during training. Taking a closer look at each category, our method outperforms NOCS on the bottle, can and camera categories. We hypothesize that the complex textures of cans and bottles is problematic for the regression of the NOCS features, and the complex geometry of cameras poses a challenge for ICP. Fig.~\ref{fig:exp_nocs_visual} shows qualitative results. It can be seen that our method produces more accurate results than NOCS, especially for geometrically complex objects. This may also provide an explanation for the narrow performance advantage of NOCS on bowl and laptop; both object types have many planar regions which favor the point-to-plane metric used in ICP. 

Finally, our method significantly outperforms the discriminative RGB-based baseline which directly regresses 3D orientation from images. This large gap is partially due to the sensitivity of discriminative methods towards distribution shifts and domain gaps between the training and test data (e.g., instances, lighting, occlusion and segmentation imperfections) and proves the generalization power of generative models for pose estimation.

\subsection{Component Analysis}

\subsubsection{Image Generation.} 
\begin{figure}
    \centering
    \includegraphics[width=\textwidth]{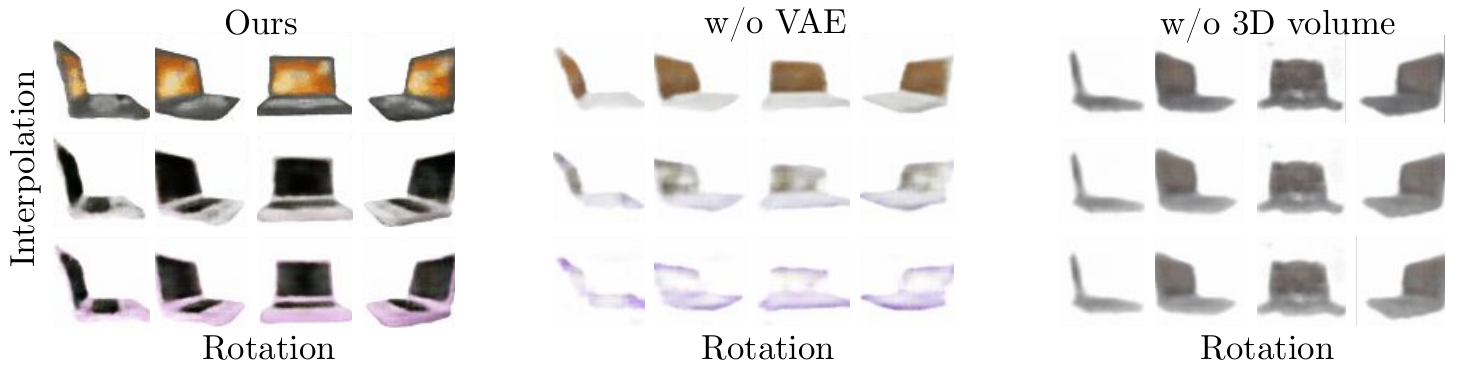}
    \caption{\textbf{Image generated with different network architectures.} Our method can generate novel images in diverse shape and appearances and in desired poses. Without using KL divergence (w/o VAE) the network cannot generate novel samples. Without using 3D structure, the network struggles to faithfully reflect rotation and hence the image quality suffers. }
    \label{fig:exp_generation}
\end{figure}

We evaluate the influence of our design choices in the image generation network on both the image generation and the object pose fitting task. First, we train a network in which the poses are directly concatenated to the latent code and then decoded to images with 2D convolutions only, the latter essentially being a standard conditional VAE (denoted by \textbf{w/o 3D}). Tab.~\ref{tab:exp_network} and Fig.~\ref{fig:exp_generation} clearly show that the lack of a 3D latent space leads to poor image generation results and consequently the fitting fails.

We further train our network without the KL divergence term (denoted by \textbf{w/o VAE} in Tab.~\ref{tab:exp_network}). While the image generation network achieves a lower training objective without regularization, this results in a non-smooth latent space, i.e. only few samples in the latent space are used for image generation. In consequence, the fitting is negatively impacted since no informative gradient can be produced to guide the updates of the latent code. 

Finally, we study the influence of the dimension of the latent space. We train several networks with different dimensions. Tab.~\ref{tab:exp_dim} summarizes the results. Low-dimensional latent spaces are not capable of representing a large variety of objects, and inflect high image reconstruction errors and high pose fitting error. On the other extreme, we find that higher dimensionality leads to better image reconstruction quality, whereas it poses difficulty for fitting. We hypothesize that this is due the more complex structure of the latent space which is required to obtain high reconstruction fidelity. Empirically, we find that a 16-dimensional latent space leads to a good trade-off between image quality and pose accuracy.

\begin{table}[!bt]
    \centering
	\caption{\textbf{Effect of network design choices on object pose estimation}. (a) demonstrates that both the VAE training scheme and the 3D feature volume are benefical for object pose estimation. (b) shows that overly high- or low-dimensional latent space can negatively influence the pose estimation.}
	\begin{subfigure}{0.5\linewidth}
        \centering
        \caption{Network architectures.}
            \setlength{\tabcolsep}{0.2cm}
            \begin{tabular}{@{}cccc@{}}
                \toprule
                    & L1   & $AP_{10}$ & $AP_{60}$ \\ \midrule
                w/o 3D & 0.109            & 1.8\%    & 90.2\%          \\
                w/o VAE & \textbf{0.049}            & 12.3\%      & 92.7\%        \\
                Ours   & 0.056             & \textbf{16.5\%}      & \textbf{97.1\%}        \\ 
                \bottomrule
            \end{tabular}
		\label{tab:exp_network}
    \end{subfigure}
    \hfill
	\begin{subfigure}{0.45\linewidth}
        \centering
        \caption{Latent space dimension.} 
            \setlength{\tabcolsep}{0.2cm}
            \begin{tabular}{@{}cccc@{}}
                \toprule
                       & L1   & $AP_{10}$ & $AP_{60}$\\ \midrule
                4      & 0.082            & 9.1\%      & 96.6\%         \\
                16     & 0.056     & \textbf{16.5\%}      & \textbf{97.1\%}         \\
                128    & \textbf{0.041}            & 11.4\%      & 96.0\%  \\
                \bottomrule
            \end{tabular}
		\label{tab:exp_dim}
	\end{subfigure}\\
\end{table}

\subsubsection{Optimization.} We study the hyper-parameters for optimization. We plot the error evolution of our energy function and the rotation error at different iterations. As evident from Tab.~\ref{tab:iterations}, with decreasing loss, the rotational error also decreases, demonstrating that our energy function provides a meaningful signal for pose estimation. In practice, our method converges in less than 50 iterations in most cases. The fitting progress is also visualized qualitatively in the figure in Tab.~\ref{tab:iterations}. The initial pose is significantly different from the target as shown in the inset. The initial appearance, which is obtained via the encoder, is close to the target appearance but still displays noticeable differences, e.g. the touchpad. Our method jointly refines both the pose and appearance. During the first few iterations, the method mainly focuses on adjusting the pose since the pose has a larger influence on the error than the appearance. In the last few iterations the method mainly focuses on fine-tuning the appearance to adapt to the image. 

\begin{table}[!bt]
    \centering
    \caption{\textbf{Ablation study of energy functions.} As shown in (a), perceptual loss outperforms other potential error functions. (b) shows that the rotational error decreases with decreasing energy value. }
	\begin{subfigure}[t]{0.35\linewidth}
        \centering
        \caption{Effect of energy functions.} 
            \setlength{\tabcolsep}{0.2cm}
            \begin{tabular}{@{}cccc@{}}
                \toprule
                            & $AP_{10}$ & $AP_{60}$ \\ \midrule
                    L1 & 14.3\%    & 85.3\%          \\
                    L2 & 13.2\%      & 88.3\%        \\
                    SSIM & 15.8\%      & 92.5\%        \\
                    w/o reg & 16.2\%      & 95.9\%        \\
                    Ours   & \textbf{16.5\%}      & \textbf{97.1\%}        \\ \bottomrule
            \end{tabular}
		\label{tab:loss}
    \end{subfigure}
    \hfill
	\begin{subfigure}[t]{0.6\linewidth}
        \centering
        \caption{Loss and pose error along iterations.} 
            \includegraphics{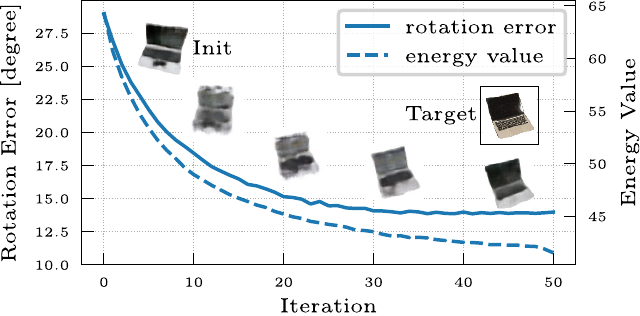}
		\label{tab:iterations}
	\end{subfigure}\\
\end{table}

We also evaluate other potential choices for the energy function to minimize, including the mean absolute error (L1), the mean squared error (L2) and the Structural Similarity Index Metric (SSIM) \cite{wang2004image}. These error functions do not perform as well as the perceptual loss as shown in Tab.~\ref{tab:loss}. This is likely due to the fact that the perceptual loss encourages semantic alignment rather than pixel-wise alignment. Thus it produces results that are globally aligned instead of focusing on local regions.  We also conduct experiments in fitting without the regularization term (\textbf{w/o reg}). This leads to unrealistic samples that minimizes the loss in an undesired way, resulting in a performance decrease compared to our full model.
\subsection{Robustness}
\begin{figure}[t]
    \includegraphics[width=\textwidth]{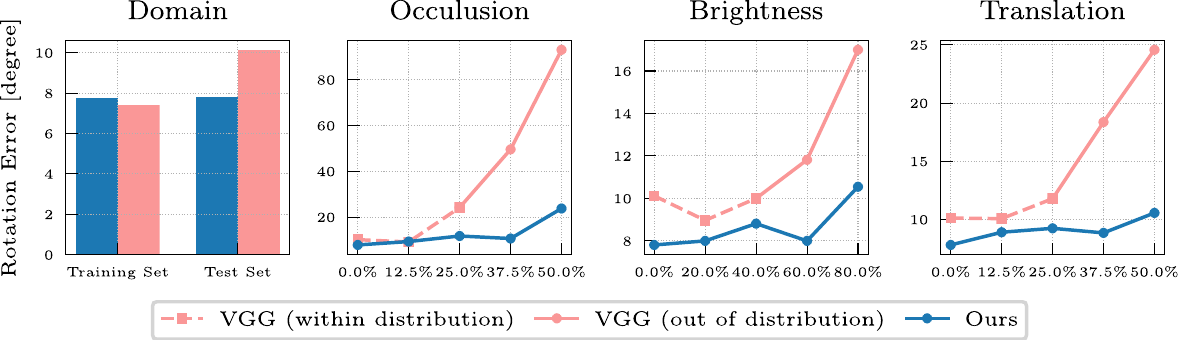}
    \caption{\textbf{Study of robustness.} We study the influence of object instances, occlusion, brightness, and translation on the accuracy of rotation estimation. Compared to the VGG-based regression baseline, ours retains low error under challenging conditions even without any data augmentation at training time.}
    \label{fig:exp_robustness}
\end{figure}
In order to study the robustness of our method to varying factors causing domain shift, we evaluate our method in a controlled simulation environment. We train our network using the ``laptop'' category of the synthetic ShapeNet dataset and test on unseen synthetic instances. We mainly study three factors that often differ between real and simulated data, namely lighting, occlusion and offsets in bounding box detection. At test time, we vary one of these factors at a time and evaluate the average error in terms of orientation estimation. We modify the brightness of the target image for lighting, remove certain areas of the image to simulate occlusion and translate the image in 2D to emulate inaccurate 2D detection.
For comparison, we train a VGG16 network to regress rotation angles from images using the same training data.
It is well known that discriminative approaches are more susceptible to overfitting which often results in worse generalization performance compared to generative approaches.
We verify this finding for the category-level pose estimation task.
We also randomly vary the three factors already when training the discriminative approach, but only to a limited degree (up to 20\% for occlusion, 40\% for lighting and 25\% for translation). At test time, we test the network on samples both within and beyond the training variations.
Note that our method \textit{never} sees augmented images during training.

Fig.~\ref{fig:exp_robustness} illustrates the results of this experiment.  First, we observe that our generative approach is less sensitive to the gap between training and test instances which is a crucial design goal of our approach in order to deal with unseen objects. When varying the three factors, the discriminative model exhibits significant performance variations especially when the factor exceeds the variations in the training distribution. In contrast, our method exhibits less performance variation which demonstrates the robustness of our method. 

\section{Conclusion}
In this paper, we propose a novel solution to category-level object pose estimation. 
We combine a gradient-based fitting procedure with a parametric neural image synthesis model that is capable of implicitly representing the appearance, shape and pose of entire object categories, thus avoiding the need for instance-specific 3D CAD models at test time. We show that this  approach reaches performance on par with and sometimes even outperforms a strong RGB-D baseline.
While our focus lies on rigid objects, extending our method to handle non-rigid or partially rigid objects is an interesting direction for future work. 

\noindent\textbf{Acknowledgement.} This research was partially supported by the Max Planck ETH Center for Learning Systems and a research gift from NVIDIA. %

\newpage
\bibliographystyle{splncs04}
\bibliography{main}

\end{document}